# Behavioral Modeling for Churn Prediction:

## Early Indicators and Accurate Predictors of Custom Defection and Loyalty

Muhammad Raza Khan[1], Joshua Manoj[2], Anikate Singh[3], Joshua Blumenstock[4]
[1]*Information School, University of Washington, Seattle, WA, USA*
Email: [1]mraza@uw.edu, [2]joshuacm@uw.edu, [3]aniksing@uw.edu, [4]joshblum@uw.edu

***Abstract*— Churn prediction, or the task of identifying customers who are likely to discontinue use of a service, is an important and lucrative concern of firms in many different industries. As these firms collect an increasing amount of large-scale, heterogeneous data on the characteristics and behaviors of customers, new methods become possible for predicting churn. In this paper, we present a unified analytic framework for detecting the early warning signs of churn, and assigning a "Churn Score" to each customer that indicates the likelihood that the particular individual will churn within a predefined amount of time. This framework employs a brute force approach to feature engineering, then winnows the set of relevant attributes via feature selection, before feeding the final feature-set into a suite of supervised learning algorithms. Using several terabytes of data from a large mobile phone network, our method identifies several intuitive - and a few surprising - early warning signs of churn, and our best model predicts whether a subscriber will churn with 89.4% accuracy.**

***Keywords*— Churn; machine learning; supervised learning; data science; call detail records.**

## I. INTRODUCTION

A common expression in industry is that it costs five times more to acquire a new customer than it does to retain an existing one. In sectors ranging from finance and insurance to cable service and internet dating, significant sums of money are spent in reducing customer attrition. To prevent such attrition (*churn*) it is critical to be able to identify the early warning signs of churn.

In this paper, we describe a data-centric and machine learning based framework for churn prediction. The framework has two primary components. The first is designed to identify early indicators of churn through a semi-supervised, brute force approach to feature engineering. The second component in our framework is designed to construct a "Churn Score" that assigns a probability to each customer indicating the predicted likelihood that the particular subscriber will churn within a predefined period of time. Whereas the focus of the first stage is interpretability, in the sense that the behavioral traits identified can be easily understood, the focus of this second stage is to create an accurate predictive metric that could, for instance, be used by a marketing department to target promotions to customers who are on the brink of defecting to a competitor. To this end, we utilize a "kitchen-sink" approach to supervised learning in which the full set of thousands of features are passed through basic classifiers, with parameters selected through cross-validation, in order to optimize standard metrics of predictive performance. While the models themselves are complex, the Churn Scores they produce are highly accurate.

We test and calibrate this framework on a large dataset from a mobile phone operator in South Asia. Starting with a raw dataset of several billion transactions, spanning roughly ten million prepaid mobile phone subscribers over a period of multiple years, we extract a calibration dataset consisting of all network-based communication for roughly 100,000 subscribers over 6 months. On this dataset, where the natural churn rate during our evaluation period is roughly 24%, our method is able to predict customer churn with just under 90% accuracy.

The remainder of the paper is organized as follows: in the following section, we discuss related research on churn prediction. Section III describes the data and methods in greater detail. Results are presented in section IV, and we conclude with a discussion of the strengths and weaknesses of our approach in section V.

## II. RELATED WORK

Understanding why customers terminate relationships has been a focus of marketing research for several decades[1]. In recent years, as data on customer activities and characteristics becomes increasingly available to companies, more sophisticated metrics have evolved to describe customer behavior and better understand how behavioral traits can be linked to customer retention and firm performance[2]. Spurred in part by a competition associated with the ACM SIGKDD Conference on Knowledge Discovery and Data Mining, churn prediction has received recent attention from the applied machine learning community. These approaches have tested a battery of models including expert systems [3], support vector machines [4], and bagging and boosting [5], to name just a few. They further vary in terms of the approach to the data, with some focusing on customer profiles and features[6], and others concerned primarily with the importance of social ties and social structure [7], [8], [9].

Zhang *et al.*[10] provide a recent overview of the different types of subscriber attributes used to model and predict customer churn in prior work, and Verbeke *et al.* [11] benchmark several classification techniques for prediction. Neslin *et al.* [12] discusses the importance of different methods for predicting churn in the context of a public tournament between 33 different competitors. In these and related studies, the behavioral traits are pre-computed -- for instance in the contest described by [12], each contestant was given a curated dataset with 171 predictor variables. By contrast, a focal point of our study is on the process of generating these predictor variables from the raw transactional records.

The empirical setting for our study -- a prepaid mobile network in South Asia -- is further distinct from almost all previous work on churn prediction in two ways. First, in the network we study, churn is a relatively common event. We observe an approximate churn rate of 25% every two months, which is an order of magnitude higher than the "rare-event" churn discussed by most previous research (e.g. [5]). Second, the vast majority of studies of churn in telecommunications focus on post-paid network where churn is equivalent to the cancellation of a contract. By contrast, our empirical setting is one of pre-paid accounts, where churn simply means that the subscriber discontinues use of the network, and must be inferred by a prolonged period of (passive) inactivity[1].

## III. DATA PRE-PROCESSING AND METHODS

We study customer churn in the context of a mobile phone use in South Asia, using data from one of the largest mobile network operators in the country[2]. The operator has roughly ten million active subscribers, and has provided us with access to an anonymized database containing several years of transactional communication histories. The transactions on which we focus are the Call Detail Records (CDR), simple metadata logs that record entries for every event that transpires on the network. Thus, for every call or text message (SMS) received, we observe the anonymized identity of the initiating and receiving party, a date and timestamp, approximate physical location of the two parties, and additional metadata describing the activity.

An overview of the framework we develop, which allows us transform the raw transactional records into a set of early warning signs of churn, as well as a subscriber-specific Churn Score, is shown in Figure 1.

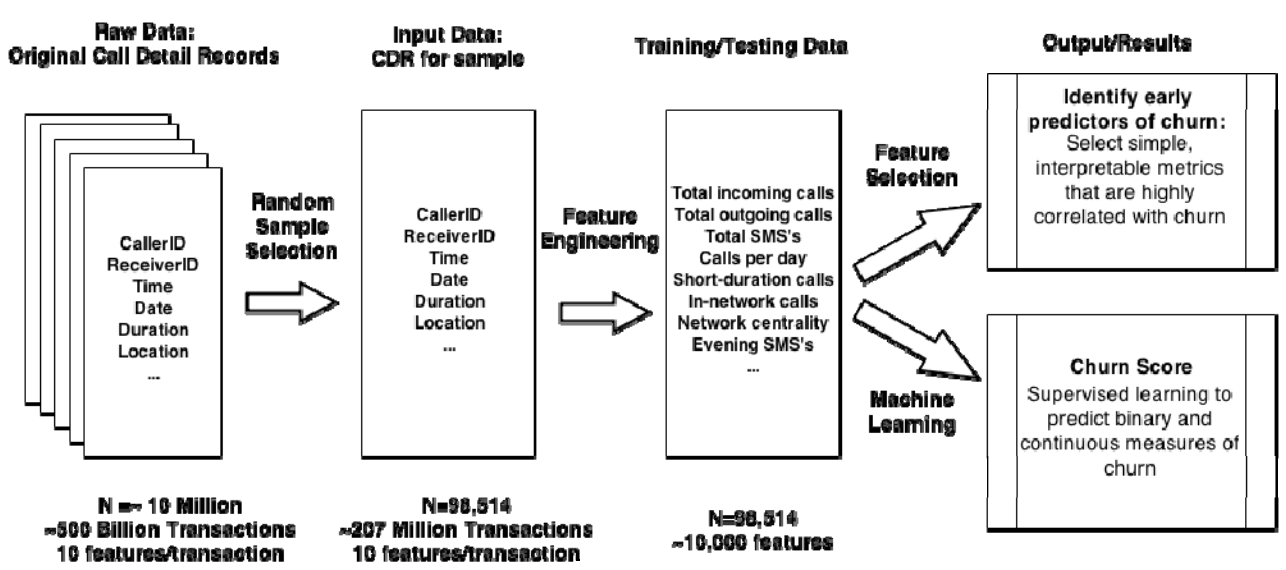

*Figure 1 Overview of our Approach*

We begin by randomly selecting a subsample of roughly 100,000 subscribers from the full mobile phone subscriber base, and extract all transactions in which they are involved. Basic summary statistics for our random sample are given in Table 1.

*Table 1 Summary Statistics*

| ***Property*** | ***Value*** |
|---|---|
| Number of Unique Subscribers | 98,514 |
| Total Number of Days | 183 |
| Total Number of Calls | 69,804,577 |
| Total Number of SMS | 137,174,182 |
| Average (Standard Deviation) of Calls per subscriber | 20 (317) |
| Average (Standard Deviation) of SMS per subscriber | 386 (1729) |

For this subset of subscribers, we then generate a large number of aggregated metrics that describe a wide range of inferred behavioral characteristics (*Feature Engineering*). The resultant dataset, which we use for training and testing the core algorithms, is a balanced rectangular matrix containing approximately 10,000 features for approximately 100,000 individuals. With this matrix, we then separately isolate the handful of metrics that are most predictive of churn (*Feature Selection*), and develop a Churn Score that indicates the likelihood that a subscriber will churn in a 3-month period (*Machine Learning*). Each of these steps is described below.

### A. *Feature Engineering*

In constructing a set of behavioral metrics from the raw transactional data, we employ a data-centric approach that is intended to minimize the role of the analyst in determining which features are relevant to the task of modeling customer churn. To this end, we use a combinatoric "brute force" technique that defines a feature along $K$ different axes (where the $i^{th}$ axis is of dimension $D_i$), then creates all possible combinations of such dimensions. This results in a total of $\prod_{i=1}^{K} D_i$ unique features.

In our case, we use eight different axes, each of which has between two and seven dimensions. A schematic of the feature space covered by these axes is given in Figure 2. Most of these axes are intuitive: we divide communication by total activity, which captures the total number of communication events of a certain type, and by the degree, which indicates the number of unique alters with which a given ego communicates. We further divide by the type of communication (call vs. SMS), the time of day, the direction of communication, type of other party number. Additionally, we segment features based on the total amount of activity over the entire observation window, as well as by the trend in activity from month to month.

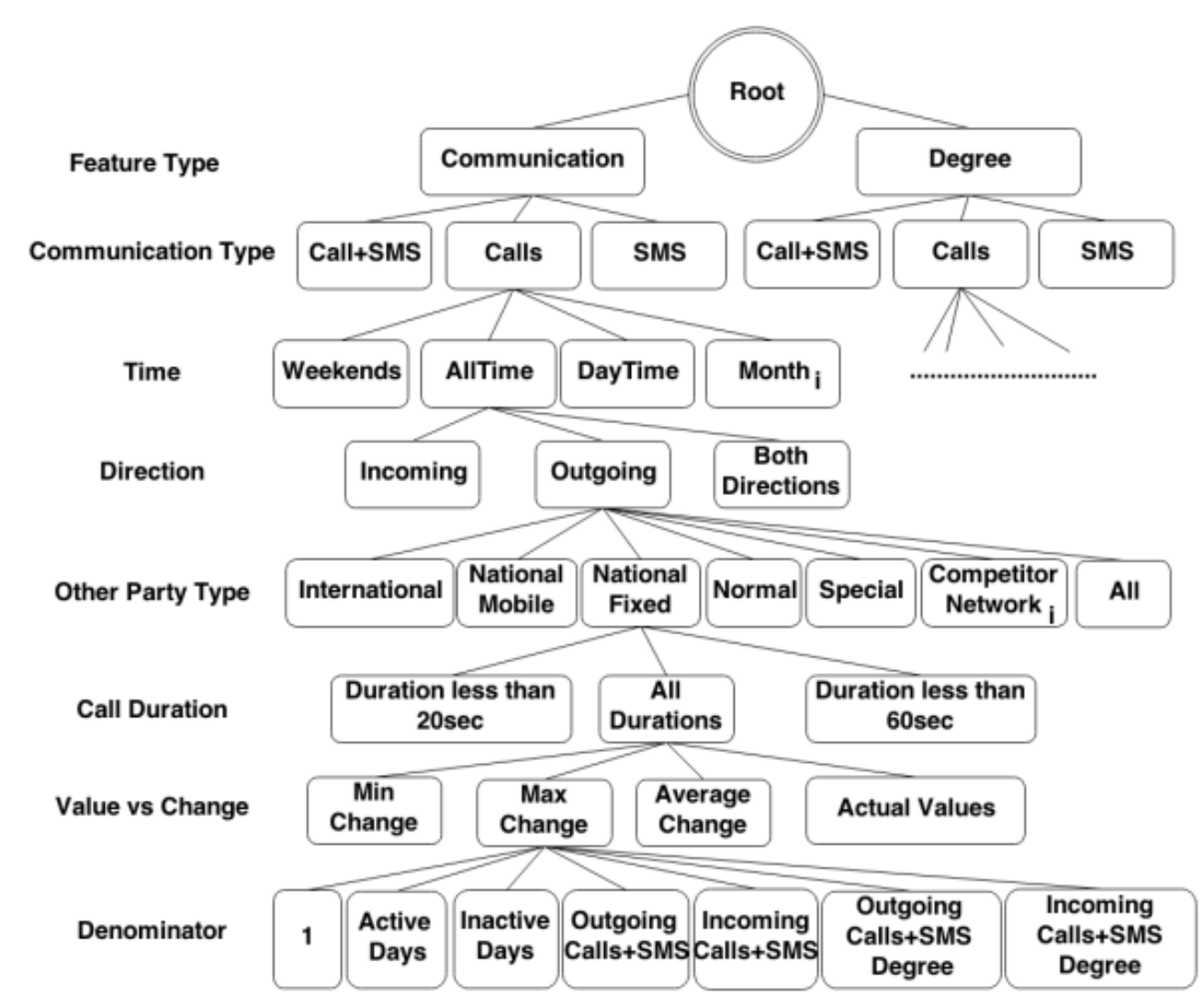

*Figure 2 Tree used for combinatoric feature generation*

[1] Dasgupta *et al.* [7] provide a noteworthy exception, focusing on a prepaid operator with annual churn rates between 50 and 70 percent.

[2] To protect the competitive interests of our data partner, specific information on the context and subscriber base is kept confidential.

## B. *Feature Selection*

Given the large number of features generated through the process described above, we then employ standard methods of feature selection that determine which features are most correlated with customer churn. We do this in two ways: We first use compute a t-test separately for each individual feature, which indicates the extent to which that single feature can accurately differentiate between people who churn and people remain active. Separately, we use a tree-based method for feature selection that allows us to estimate the conditional ability of each additional feature to improve the overall accuracy of a joint classifier [13].

## C. *Churn Prediction and Churn Scoring*

The process of feature selection describe above is useful in producing an interpretable list of metrics that are correlated with customer churn. Our second goal is to use these features to predict churn. Specifically, we seek to assign each subscriber a Churn Score between zero and one that indicates the likelihood that the subscriber will churn. To this end, we set up a simple supervised learning experiment where we divide our data into a training period and an evaluation period, derive features using the above process during the training period, and use these features to predict churn during the evaluation period. Using k-fold cross-validation, we then train several different supervised learning algorithms on randomized subsets of individuals. The specific algorithms we test are linear and logistic regression, Support Vector Machines, K-Nearest Neighbors, Random Forests, and AdaBoost (with a decision tree classifier)[3]. A common assumption made in industry regarding churn is that a customer is considered lost after 2-3 months of continuous inactivity; thus, in some of the experiments described below we construct a binary measure of churn that approximates this condition.

# IV. RESULTS

We set up our experiment using a 6-month period of data, (first four months training and last 2 months testing). We quantify churn in two ways: first, as a binary condition that is true if the subscriber is completely inactive for the entire 2-month window. In total, 26% of our subscribers fit this stringent definition of churn. Second, we define a more flexible version of churn as the percentage of days on which no activity is observed. The distribution of inactivity for our sample can be seen in Figure 3.

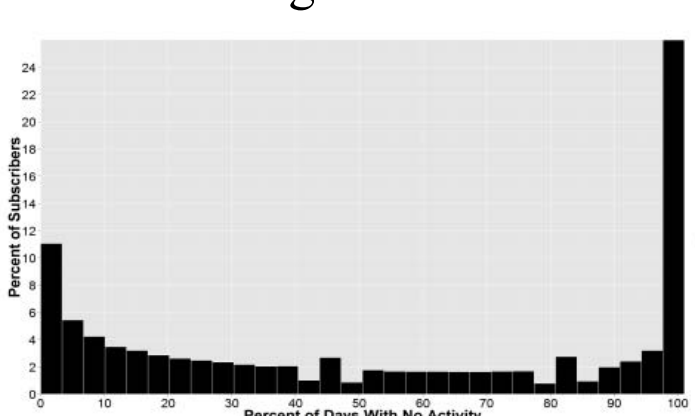

*Figure 3 Distribution of Inactivity*

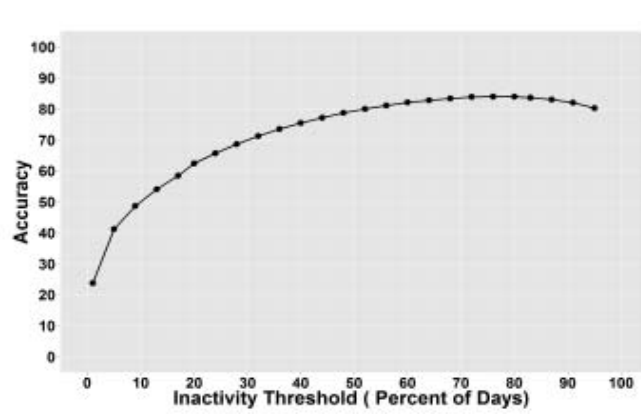

*Figure 4 Churn Prediction for a Single Linear Discriminant*

## A. *Early Predictors of Mobile Subscriber Churn*

Using the approach described in Section IV.A, we generate 12,914 unique features, each of which quantifies a different type of mobile phone-based behavior. In Panel A of Table 2, we list the 10 features that, taken in isolation, are most predictive of churn. For the ease of exposition, we focus initially on the continuous definition of churn, and indicate in the right column the $\mathbf{R^2}$ that results from separately regressing the proportion of days with no activity on each feature. Unsurprisingly, "Percent of inactive days" is highly predictive of future churn

The set of features listed in Table 2 are all unconditionally highly correlated with churn, these features are also correlated with one another. Thus, while the "Percent of inactive days" feature may be the best single linear discriminant between churners and non-churners, it is not necessarily the case that the ensemble of 10 features in Panel A will be the best *joint* predictor of churn. Thus, in Panel B of Table 2, we provide the 10 features that are, taken together, the best joint predictors of subscriber churn. For comparison with Panel A, we also list the $\mathbf{R^2}$ from the unconditional (univariate) regression for each of the features.

*Table 2 Top Predictors of Churn*

| Rank | Feature | $R^2$ |
|---|---|---|
| | *Panel A: Tested individually* | |
| 1. | Percent of inactive days (during training period) | 0.66 |
| 2. | Maximum monthly Δ in incoming calls / Total incoming calls | 0.44 |
| 3. | Maximum monthly Δ in incoming calls / Total outgoing calls | 0.42 |
| 4. | Outbound network degree (most recent month) | 0.35 |
| 5. | Incoming text messages received from competitor's network | 0.33 |
| 6. | Average number of calls to Information Portal per active day | 0.33 |
| 7. | Unique weekend contacts per active day | 0.33 |
| 8. | Average daily text messages received from competitor's network | 0.33 |
| 9. | Number of Inactive Days in the First Month of the Training Period | 0.31 |
| 10. | Daytime degree (voice calls) | 0.25 |
| | *Panel B: Tested jointly* | |
| 1. | Outgoing degree (most recent month) | 0.35 |
| 2. | Outgoing degree (first month) / SMS Degree | 0.09 |
| 3. | Incoming degree (second month) / Total incoming calls | 0.14 |
| 4. | SMS to Mobile Money Service / Total days of inactivity | 0.01 |
| 5. | Short-duration calls (first month) / Total incoming calls | 0.24 |
| 6. | Incoming calls / Incoming events | 0.12 |
| 7. | Calls to Mobile Money servce (first month) / Total days of inactivity | 0.32 |
| 8. | Outgoing events (first month) / Call degree | 0.22 |
| 9. | Total incoming int'l SMS (weekends) / Incoming degree | 0.18 |
| 10. | Total outgoing degree (first month) / Call degree | 0.22 |

The conditionally predictive features in Panel B of Table 2 tend to be micro-aggregates; the final set is thus less redundant and less mechanically correlated.

## B. *Predicting Churn*

To evaluate the performance of our predictive framework, we first construct a simple and intuitive baseline that builds on the results of Section V.A. Namely, we select the single best unconditional predictor of churn, "Percent of inactive days", and build a linear discriminant model based on that feature. We have tested all possible threshold values of this metric to determine the value that most accurately separates churners from non-churners, where we now define churn as a binary indicator that takes the value one for churners and zero for non-churners. As shown in Figure 4, a maximum accuracy of 83.9% for this metric is obtained at a threshold of 76. In other words, when we classify as churners all subscribers who were inactive on more than 76 percent of days during the training period, our prediction is correct in 83.9% percent of cases. The good performance of this simple linear discriminant is due to the fact that we have an unbalanced sample, in the sense that 76.6 % of our sample do not churn, and a very simple model that just predicted the majority class (i.e. "not churn") for all subscribers would achieve 76.6% accuracy.

[3] See [11] for a more thorough exposition of the impact of different learning algorithms on churn prediction accuracy.

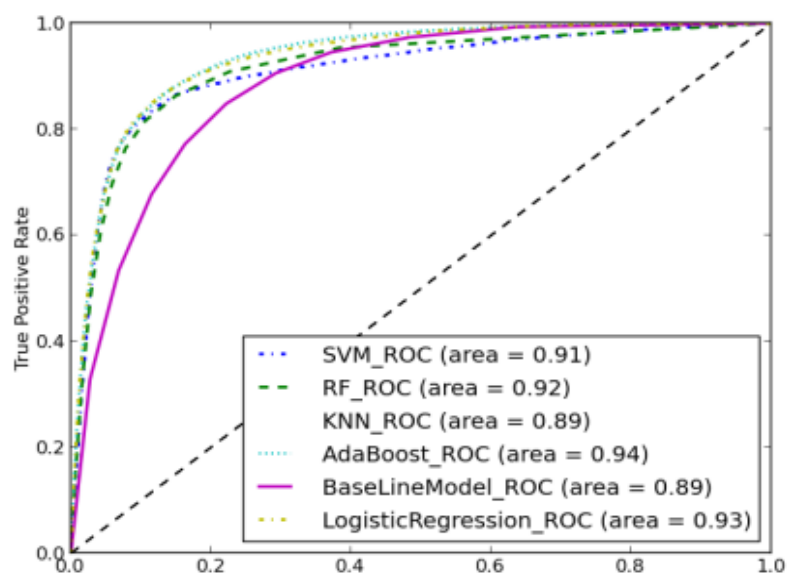

*Figure 5 ROC Curves*

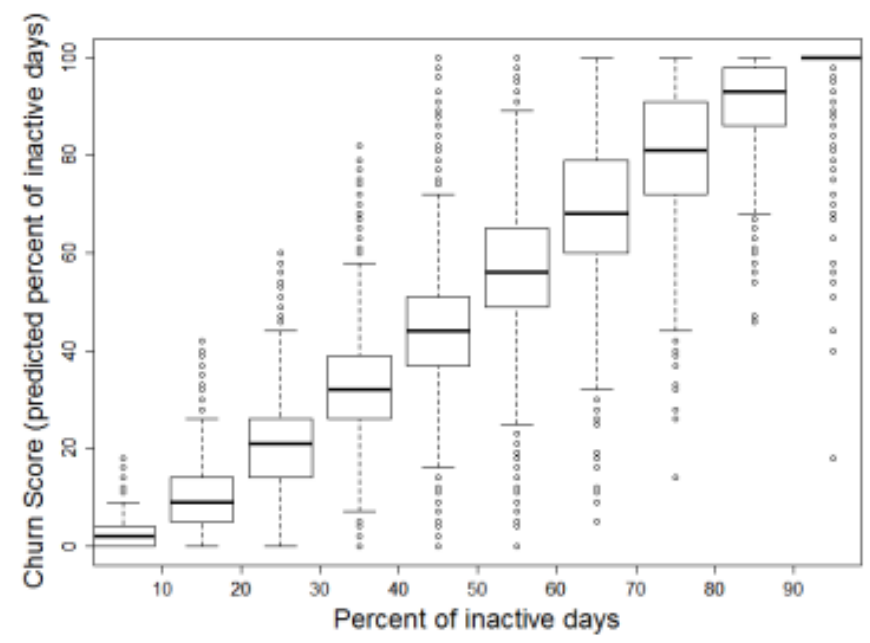

*Figure 6 Actual vs Predicted % of Inactive Days*

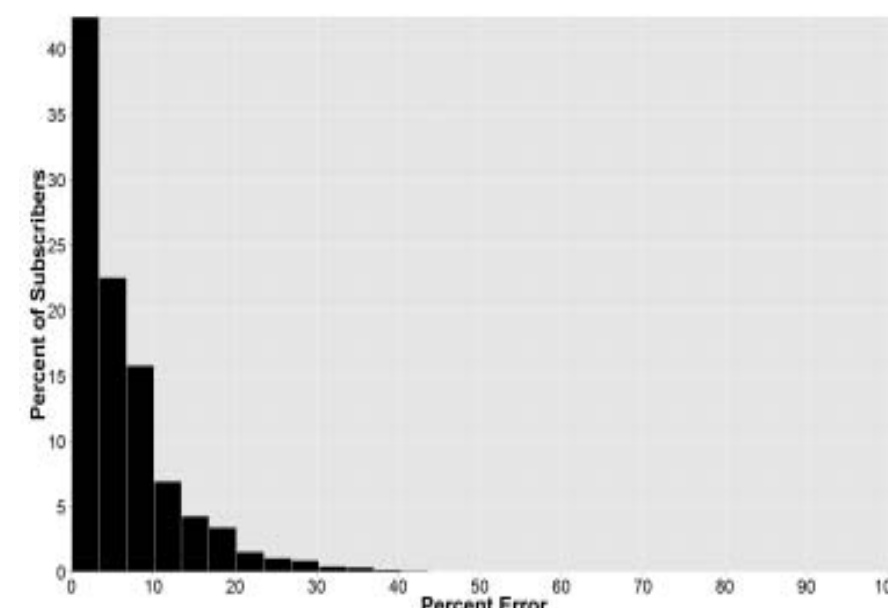

*Figure 7 Distribution of Prediction Errors*

Depending on the algorithm used to predict churn, we achieve accuracy rates of roughly 88.5-89.5% (Table 3). This represents a modest improvement of roughly 6% over the single-feature baseline, or approximately 14% over the majority-class baseline. A variety of performance characteristics of each model, as well as the linear discriminant baseline, are given in Table 3; the ROC curves for these models is presented in Figure 5. For each model, we use the set of 100 features selected via the bagging approach described in Section V.A. Similar results are obtained when considering churn as a continuous measure of inactivity. This metric, which we denote as the "Churn Score", indicates the percentage of days in which the subscriber is predicted to be inactive. Individuals with high Churn Scores are the likely churners. Figures 6 and 7 show prediction accuracy, measured as the absolute difference between the predicted percentage of inactive days and the actual percentage of inactive days, measured for each subscriber during the evaluation period.

## V. Discussion and Conclusion

In this paper, we have presented a simple framework for detecting the early warning signs of churn, and for developing a Churn Score to identify subscribers who are likely to end their relationship with the company. The approach uses a brute force approach to feature engineering that generates a large number of overlapping features from customer transaction logs, then uses two related techniques to identify the features and metrics that are most predictive of customer churn. These features are then fed into a series of supervised learning algorithms that can accurately predict subscriber churn. Testing this approach on several terabytes of data from a South Asian mobile phone operator, we highlight a few of the early predictors of churn, and show that relatively simple models predict churn with close to 90 percent accuracy.

While these initial empirical results are promising, we see the primary contribution of this paper being the description of a systematic framework that can be used to generate interpretable features and predict customer outcomes. Several of the modelling assumptions we have made, such as the axes and dimensions used to generate features, are quite arbitrary and careful design of these behavioral metrics may yield more intuitive predictors and more accurate predictions. In future work, we are interested in systematic exploration of the feature space requiring least involvement from the analysts.

*Table 3 Predictive Performance of Classification Algorithms*

| Model | Accuracy | Precision | Recall | F-Score | AUC |
|---|---|---|---|---|---|
| SVM | 89.4 | 0.89 | 0.89 | 0.89 | 0.91 |
| Random Forest | 88.4 | 0.88 | 0.88 | 0.88 | 0.92 |
| KNN | 88.2 | 0.88 | 0.88 | 0.89 | 0.89 |
| AdaBoost | 89.2 | 0.89 | 0.89 | 0.89 | 0.94 |
| Logistic Regression | 89.3 | 0.89 | 0.89 | 0.89 | 0.93 |
| Baseline Model | 83.9 | 0.83 | 0.84 | 0.83 | 0.89 |